\documentclass[runningheads]{llncs}
\usepackage{makeidx}
\usepackage{graphicx}
\usepackage{caption}
\usepackage{subcaption}
\usepackage{booktabs}
\usepackage{amsmath,amssymb} 
\usepackage{color}
\usepackage{authblk}
\usepackage[hidelinks]{hyperref}
\usepackage{url}

\begin{document}
	\pagestyle{headings}
	\mainmatter
	\title{Multi-label Object Attribute Classification using a Convolutional Neural Network}
	\titlerunning{Multi-label Object Attribute Classification using a ConvNet}
	\author{Soubarna Banik, Mikko Lauri, Simone Frintrop\vspace{-0.27cm}}
	
	\institute{\href{mailto:4banik@informatik.uni-hamburg.de}{\url{ \string{4banik,lauri,frintrop  \string}@informatik.uni-hamburg.de}} \vspace{0.25cm} \linebreak Department of Informatics, Universit{\"a}t Hamburg}
	
	\maketitle

	\begin{abstract}
		Objects of different classes can be described using a limited number of attributes such as color, shape, pattern, and texture. 
		Learning to detect object attributes instead of only detecting objects can be helpful in dealing with a priori unknown objects. 
		With this inspiration, a deep convolutional neural network for low-level object attribute classification, called the Deep Attribute Network (DAN), is proposed. 
		Since object features are implicitly learned by object recognition networks, one such existing network is modified and fine-tuned for developing DAN. 
		The performance of DAN is evaluated on the ImageNet Attribute and a-Pascal datasets. 
		Experiments show that in comparison with state-of-the-art methods, the proposed model achieves better results.
	\end{abstract}

	\section{Introduction}
	\label{sec:Intro}
	In recent years, the problem of attribute classification has attracted substantial interest within the computer vision community.
    Attributes can be viewed as descriptive properties of objects \cite{Farhadi2009}.
    Different types of object attributes have been explored in research: low-level visual adjectives (such as \textit{color}, \textit{shape}) \cite{Russakovsky2012}, inherent object characteristic (such as \textit{material}) \cite{Farhadi2009} or high-level object components (such as \textit{has-tail}, \textit{wears-sunglasses}) \cite{Chung2012,Farhadi2009}.
Recognizing object attributes has useful applications. 
Objects, though belonging to different classes, share low-level attributes such as color, shape, material. 
Attributes help to compare objects and categorize them.
Unknown objects can be described by means of a few reference attributes.
Their similarity/dissimilarity with known objects may help to infer some of their other characteristics.
For example, a robot which is familiar with an orange and knows how to grasp it, can deduce how to do the same for an unknown but similarly shaped object, if it learns the shape attribute \textit{round}.
Another important application of attribute learning is attribute-level object localization. 
When we do not have exact information about an object's class but only have partial knowledge about it, such as its shape or color, an attribute-level search can help in localizing the desired target with the help of the descriptive attributes.
For known objects, the additional attribute information can help in further reducing the search space.
Only attributes which can be described using words, i.e., semantic attributes~\cite{Russakovsky2012}, are considered in this work.
Traditional approaches \cite{cimpoi2015deep,hu2015vehicle,Russakovsky2012} use handcrafted features for low-level attribute classification.
In a few recent approaches, learned features are used instead of fixed handcrafted features \cite{Chung2012,Wang2016}.
However, the types of attributes learned are limited to complex high-level attributes like human attributes (\textit{gender}, \textit{dress type}, \textit{facial expression}, \textit{wears-hat}) or object-part attributes (\textit{wheel}, \textit{side-mirror}, \textit{window}).
These attributes are very specific to certain objects and are not universally applicable.
Learning generic low-level attributes like \textit{color}, \textit{shape}, \textit{pattern}, and \textit{texture} can be beneficial as they can describe a wide range of objects. 
For this paper, we consider learning low-level object attributes.

Object attributes are often correlated.
For example, the attribute \textit{mustache} usually co-exists with the attribute \textit{male}.
This correlation is treated in different ways in research.
In \cite{Russakovsky2012} \cite{zhang2014panda}, the correlation between attributes is ignored and the attributes are learned independently of each other.
On the other hand, in \cite{Wang2016} \cite{huang2015learning}, attributes are jointly learned by following a multi-task learning method and by explicitly modeling the correlation between the attributes.
Prediction of one attribute will increase the probability of the correlated attributes in this case.
However, low-level attributes like colors and shapes are independent of each other.
Colors such as \textit{green} and \textit{brown} may co-exist heavily in vegetation related images, but this correlation is not universal and may not exist in other types of images.
Embedding this correlation into the learning process will make the model more prone to overfitting the training data.

\begin{figure}[t!]
\includegraphics[width=\textwidth]{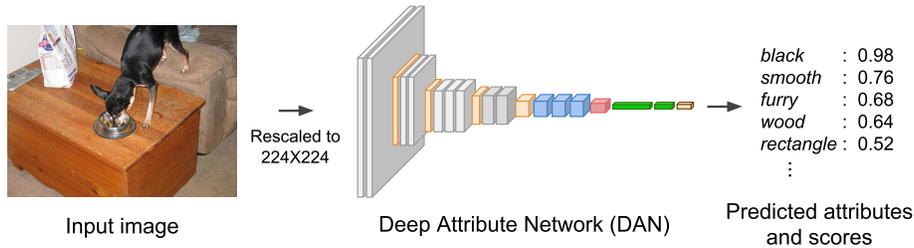}
  \caption{Overview of the proposed multi-label attribute classification system.}
  \label{fig:DAN_overview}
\end{figure}
In this paper we propose a model, named the \textit{Deep Attribute Network} (DAN) that learns low-level object attributes belonging to the attribute groups \textit{color}, \textit{shape}, \textit{pattern} and \textit{texture}.
For this purpose, an existing object classification convolutional network (ConvNet) is adapted. 
ConvNet features have been used for attribute classification in the past \cite{cimpoi2015deep,razavian2014cnn}.
However, the features are used without any customization with respect to the target task.
These features, extracted from the higher layers of the ConvNet are more specific to the objects, rather than to the object attributes.
In our approach, the layers of the base object network are finetuned for the task of object attribute classification. 
Object attribute classification is treated as a multi-label classification task and necessary adaptations are incorporated into the base network.
We learn the attributes jointly so that the model can share low-level features like edges and corners across classes.
The class label correlations are not enforced explicitly during the learning process.
Fig. \ref{fig:DAN_overview} shows an overview of the proposed attribute classification system.
To the best of our knowledge, ours is the first deep learning model that learns low-level object attributes and uses a multi-label classification approach.

Our network is trained on the ImageNet Attribute dataset~\cite{Russakovsky2012}.
For evaluation we use the ImageNet Attribute dataset \cite{Russakovsky2012} as well as the a-Pascal dataset \cite{Farhadi2009}.
We compare the performance of our model with that of state-of-the-art methods and show that the proposed model outperforms the existing approaches.
The rest of the paper is organized as follows: section \ref{sec:related_work} discusses the state-of-the-art in attribute classification.
The architecture of the proposed model is described in section \ref{sec:archi}.
Section \ref{sec:exp} discusses the performance of our model in comparison with state-of-the-art methods.
We conclude and present the future work in section \ref{sec:conclusion}.

\section{State of the art}
\label{sec:related_work}
Traditional attribute learning methods \cite{Farhadi2009,ferrari2008learning,lampert2009learning,Russakovsky2012} treat the multi-label object attribute classification problem as a set of $n$ independent binary classification problems, where $n$ is the number of target attribute classes.
These approaches use handcrafted feature descriptors, such as color and texture descriptors \cite{Russakovsky2012}.
In \cite{cimpoi2015deep,liu2015deep,razavian2014cnn}, generic feature descriptors are generated from object classification ConvNets, and then the same binary classification approach is followed. 

In a few object recognition tasks \cite{Farhadi2009,lampert2009learning}, attributes are used as an intermediate representation. Farhadi et al. \cite{Farhadi2009} shift the goal of object recognition from naming the objects to describing the objects, which helps in case of unseen object classes or classes with very few examples. Lampert et al. \cite{lampert2009learning} take a similar approach and learn the relationships between image and attributes, and between attributes and object classes. This allows the transfer of information from training classes to test classes through the intermediate attribute representation. With proper choice of attributes, their method can detect new object categories without any re-training step.

Most of the early studies on attribute classification do not consider the correlations among attributes and train each attribute classifier independently \cite{Farhadi2009,Russakovsky2012}.
Recent methods \cite{huang2015learning,Wang2016} preserve the contextual correlation between attributes by using mechanisms such as contextual Restricted Boltzmann Machines (RBM) or special graph based formulations.
In other related multi-label classification problems such as image captioning and multi-class multi-label object classification, this correlation is learned using a unified ConvNet-Recurrent Neural Network based framework \cite{wang2016cnn}.

In addition to this, there exist a few approaches, where a joint learning technique is followed with a focus on de-correlating the attributes  \cite{abdulnabi2015multi,jayaraman2014decorrelating}.
These methods encourage sharing more information among attributes belonging to the same group and less information for attributes from different groups, thereby promoting certain kinds of correlation over others.
Jayaraman et al. \cite{jayaraman2014decorrelating} implement this by weighting the feature channels differently for different attribute groups, although in a non-deep learning set up.
Abdulnabi et al. \cite{abdulnabi2015multi} propose $n$ ConvNets (for $n$ target attributes), that share a latent fully-connected layer before the output layer.
They employ regularization techniques during the training process for intra-group feature sharing and inter-group feature competition.
The provision of having an individual ConvNet for each attribute requires a large number of parameters.
Also, Abdulnabi et al. \cite{abdulnabi2015multi} follow a late fusion approach (just before the output layer), which generates redundant information in the early layers of the individual networks.

Our goal is to formulate the multi-label classification problem for low-level object attributes using a single ConvNet, where the attributes are learned jointly.
The nodes in the last fully-connected layer of a ConvNet share the hidden layers of the network.
This enables them to capture the relationship between the classes implicitly, and hence no additional mechanism is incorporated to enforce the class correlations.

\section{Architecture}
\label{sec:archi}
	
ConvNets trained on images learn simple features like Gabor filters or color blobs in the first few layers, irrespective of the training objectives \cite{yosinski2014transferable}.
These features are useful for learning our target attribute classes - \textit{color}, \textit{shape}, \textit{pattern} and \textit{texture}.
Features computed in the last layers of ConvNets are more specific to the task and the dataset at hand \cite{yosinski2014transferable}. 
 Object classes often contain instances of varied colors (e.g. white tigers, yellow tigers), patterns (e.g. dog with/without spots) or other variable features.
 This encourages the object recognition network to ignore these attributes in the later layers in order to generalize across the object class instances.
 Based on these observations, we finetune the last few layers of an object classification ConvNet (VGG-16 \cite{simonyan2014very}) in order to adapt to the attribute classification task. 
\begin{figure}[b!]
\includegraphics[width=\textwidth]{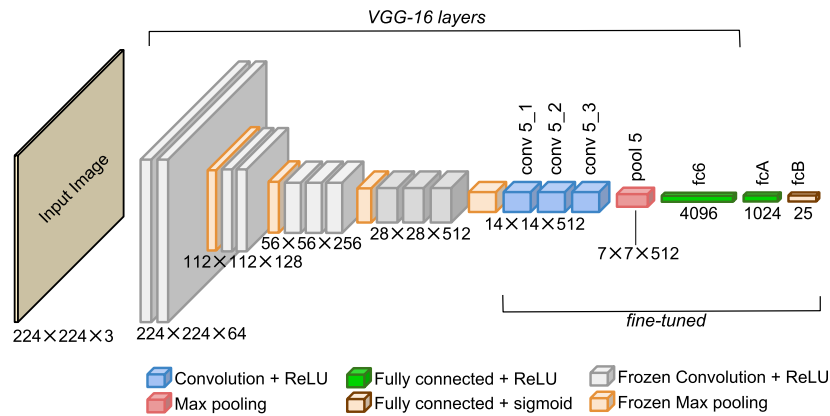}
  \caption{Architecture of the Deep Attribute Network (DAN). The network is adapted from the VGG-16 model \cite{simonyan2014very}, by replacing the last two fully-connected layers with the new layers $\textit{fcA}$ and $\textit{fcB}$. DAN is finetuned till layer $\textit{conv5\_1}$.}
  \label{fig:archi2}
\end{figure}

The proposed Deep Attribute Network (DAN) for low-level attribute classification is shown in Fig. \ref{fig:archi2}. The network is trained using transfer learning starting from the VGG-16 network, pre-trained on the ImageNet dataset \cite{russakovsky2015imagenet}. All convolutional layers and the first fully-connected layer ($\textit{fc6}$) of VGG-16 are retained in DAN. Two new fully-connected layers $\textit{fcA}$ with 1024 units and $\textit{fcB}$ with 25 units are added at the end.
The 25 units in the $\textit{fcB}$ layer correspond to the 25 target attribute classes in the ImageNet Attribute dataset \cite{Russakovsky2012} belonging to the attribute groups \textit{color}, \textit{shape}, \textit{pattern} and \textit{texture}.
All convolutional layers and fully-connected layers except $\textit{fcB}$ use ReLU activation.

Each image in the ImageNet dataset \cite{russakovsky2015imagenet} represents a single object class. Hence, VGG-16 uses softmax as the final layer to represent a categorical distribution over the mutually exclusive object classes. 
Object attributes, however, are not mutually exclusive and multiple attributes can co-occur in an object.
For example, the Swedish flag is \textit{blue}, \textit{yellow}, and \textit{rectangular} at the same time.
 Therefore, the softmax layer of VGG-16 is replaced by a sigmoid layer in DAN. 
 The last fully-connected layer $\textit{fcB}$ and the sigmoid layer are represented jointly in Fig. \ref{fig:archi2}. 
 The sigmoid layer computes the sigmoid score $\sigma(o_j)=\frac{1}{1+e^{-o_j}}$
for each element $o_j$ in layer $\textit{fcB}$, thereby converting the classifier score for each attribute class to a probability score in the range $[0,1]$. 

\paragraph{\textbf{Training}:} DAN is trained by minimizing a weighted cross-entropy loss using the stochastic gradient descent (SGD) algorithm. 
Cross-entropy is one of the popular choices as a cost function as it is non-negative, and also because the function output remains close to zero if the output score vector $\sigma(\textbf{o})$ is similar to the label vector $\textbf{y}$. 
However, the ImageNet Attribute dataset \cite{Russakovsky2012} suffers from the problem of class imbalance. As a result, the network is found to learn the class frequencies rather than the classes, when cross-entropy loss is used. To overcome this, we define the weighted cross-entropy loss $L$ as 
\begin{equation}
L(\textbf{y},\sigma(\textbf{o})) = -\sum_{k=1}^N w_k y_k \log \sigma(o_k)\,,
\label{eq:loss_wt_ce}
\end{equation}
where $k$ runs over $N$ output classes.
The weight vector $\textbf{w}$ is calculated from the labels of training samples in a batch. Weight of class $k$, $w_k$ is defined as 
\begin{equation}
w_k=1-\frac{1}{B}\sum_{j=1}^{B}y_{kj}\,,
\end{equation}
where $B$ signifies the number of images in a batch. Eq. \eqref{eq:loss_wt_ce} ensures higher loss weight for classes having more samples and low weights for classes with fewer samples. With a larger batch size and randomly shuffled data, the dynamically computed loss weight vector reflects the actual class distribution on the entire dataset. Batch size of $32$ is used for both training and test images.

The size of the input image used in DAN is the same as in VGG-16, i.e., $224\times 224$. The convolutional layers and the first fully-connected layer of the network are initialized from VGG-16, as mentioned before. The weights in the last two fully-connected layers are initialized from a Gaussian distribution with mean $0$ and standard deviation of $0.005$, and the biases are initialized to a constant value of $0.1$. The network is trained using the ImageNet Attribute dataset \cite{Russakovsky2012}.

The training is done in two phases. First, only the fully-connected layers $\textit{fc6}$, $\textit{fcA}$ and $\textit{fcB}$ are trained. 
Once the loss computed on the validation set saturates or starts to increase, the preceding block of convolutional layers (till layer $\textit{conv5\_1}$) are unfrozen and finetuned together with the fully-connected layers. 
The network is trained for $38$ epochs. 
The training is regularized by weight-decay and dropout regularization in the first two fully-connected layers. 
The base learning rate is set to $0.001$ and is decreased further by factors of $10$ when the validation-loss does not improve.
All other training parameters are set as follows - momentum to $0.9$, weight decay to $0.0005$, and dropout rate to $0.5$. 

\section{Experiments}
\label{sec:exp}
In this section, we present the quantitative evaluation of the Deep Attribute Network on the ImageNet Attribute and a-Pascal datasets. 
We implement our method using the Caffe framework~\cite{jia2014caffe}.

	\subsection{Datasets, Pre-processing and Metrics}
	\label{sec:dataset}
	
	\paragraph{\textbf{ImageNet Attribute:}} The ImageNet Attribute dataset~\cite{Russakovsky2012} contains low-level attribute annotations of objects. This dataset is a subset of the ImageNet dataset ~\cite{russakovsky2015imagenet} and consists of 9600 images from 384 synsets, each annotated with 25 attributes belonging to four categories: color (\textit{black, blue, brown, gray, green, orange, pink, red, violet, white, yellow}), shape (\textit{long, round, rectangular, square}), pattern (\textit{spotted, striped}), and texture (\textit{furry, smooth, rough, shiny, metallic, vegetation, wooden, wet}). The labels contain three values -1, +1 and 0, indicating negative, positive and ambiguous samples for which the annotators could not agree on one single attribute. 
We divide the dataset into training, validation, and test sets with 5760, 1920 and 1920 images, respectively. 
	
	\paragraph{\textbf{a-Pascal:}} The a-Pascal dataset \cite{Farhadi2009} is based on the PASCAL VOC 2008 object recognition dataset~\cite{pascal-voc-2008}. 
	The images contain objects from 20 classes and are annotated with 64 attributes, describing shape, material, and high-level object components. 
	We select three shape attributes (\textit{2D boxy, 3D boxy, round}) and five material attributes (\textit{metal, wood, furry, shiny, vegetation}) for our evaluation. 
	The shape attributes \textit{2D boxy} and \textit{3D boxy} are not present in the training dataset. 
	After manually reviewing the images, we found that both attributes are similar in appearance to the \textit{rectangle} attribute of ImageNet Attribute dataset.
	Hence, we merge these two attributes and evaluate them as the \textit{rectangle} attribute.
	The labels in this dataset are marked as 0 and 1 for negative and positive samples respectively.

	\paragraph{\textbf{Pre-processing:}} 
	The training and validation images contain object level annotations. Therefore, they are pre-processed by cropping the images to the bounding box of the object with an added margin of 10\%. 
	Additionally, the mean RGB value of the training set is subtracted from all images. 
	The label values (-1, 0,+1) in the ImageNet Attribute dataset are mapped to (0, 0.5, 1) for proper loss computation. 
	
	\paragraph{\textbf{Data augmentation:}}
	The pre-processed images are scaled to size $256\times256$. 
	At runtime, patches of size $224\times224$ are cropped randomly from the scaled training images, and from the center for the validation images. 
	The training set is further augmented by randomly flipping the images horizontally. 
	
	\paragraph{\textbf{Evaluation metrics:}} The metrics we use for evaluation are the Receiver Operator Characteristic (ROC) curve, the Precision-Recall (P-R) curve, area under the ROC curve (ROC-AUC), and the Average Precision (AP) score for the P-R curve.
In a multi-label classification setup, AP for multiple classes is reported as the \textit{mean average precision} (mAP). 
The mean can be computed in micro or macro mode~\cite{tsoumakas2009mining}. 
In micro mode, the precision and recall are calculated from the overall true/false positive/negative values across all classes.
In macro mode, the average of class-wise precision and recall are computed.
The macro-average method gives equal weight to each class, whereas the micro-average method gives equal weight to each classification decision for a sample. 
The mAP is calculated using the precision, recall values according to
$mAP = \sum_n (R_n -R_{n-1}) P_n$,
where $R_n$ and $P_n$ denote the recall and precision at $n^{th}$ threshold.

	\subsection{Attribute Classification Evaluation}
	    \begin{figure}[t!]
    \centering
    \includegraphics[width=\linewidth]{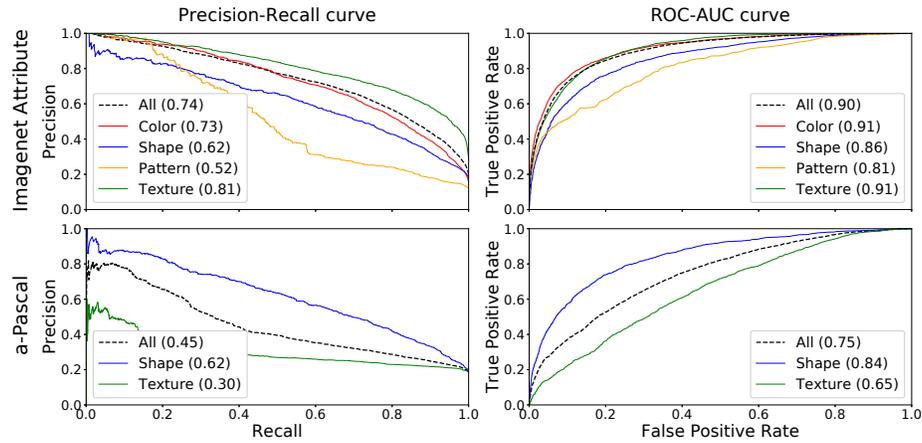}
    \caption{
    Performance of the Deep Attribute Network on the ImageNet Attribute and a-Pascal datasets. The figure shows the Precision-recall (left column) and the ROC-AUC curves (right column) for all attributes as well as for each attribute group. Within parentheses in the legends, the micro-mode mAP scores (left column) and the ROC-AUC scores (right column) are reported.
     } 
    \label{fig:classification_eval}
    \end{figure}
\begin{figure}[tb!]
\centering
\includegraphics[width=\linewidth]{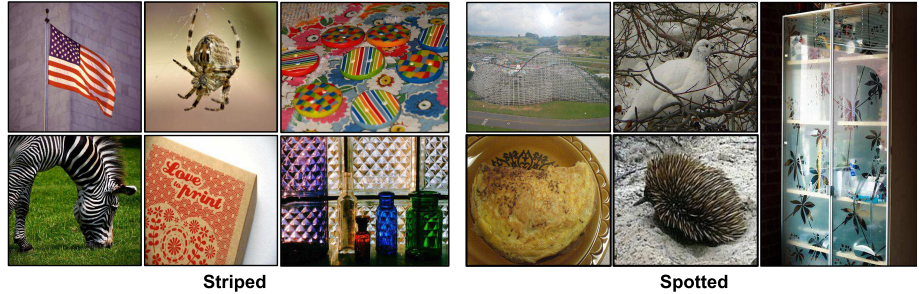}
\caption{Sample images from the class \textit{striped} and \textit{spotted} of the ImageNet Attribute dataset showing the large variety in appearance. Some images also contain incorrect labels, for example the box and the bottle in the second row of striped examples and the roller coaster and the bird in the first row of spotted examples.}
\label{fig:striped_example}
\end{figure}
\begin{figure}[t!]
\centering
\includegraphics[width=\linewidth]{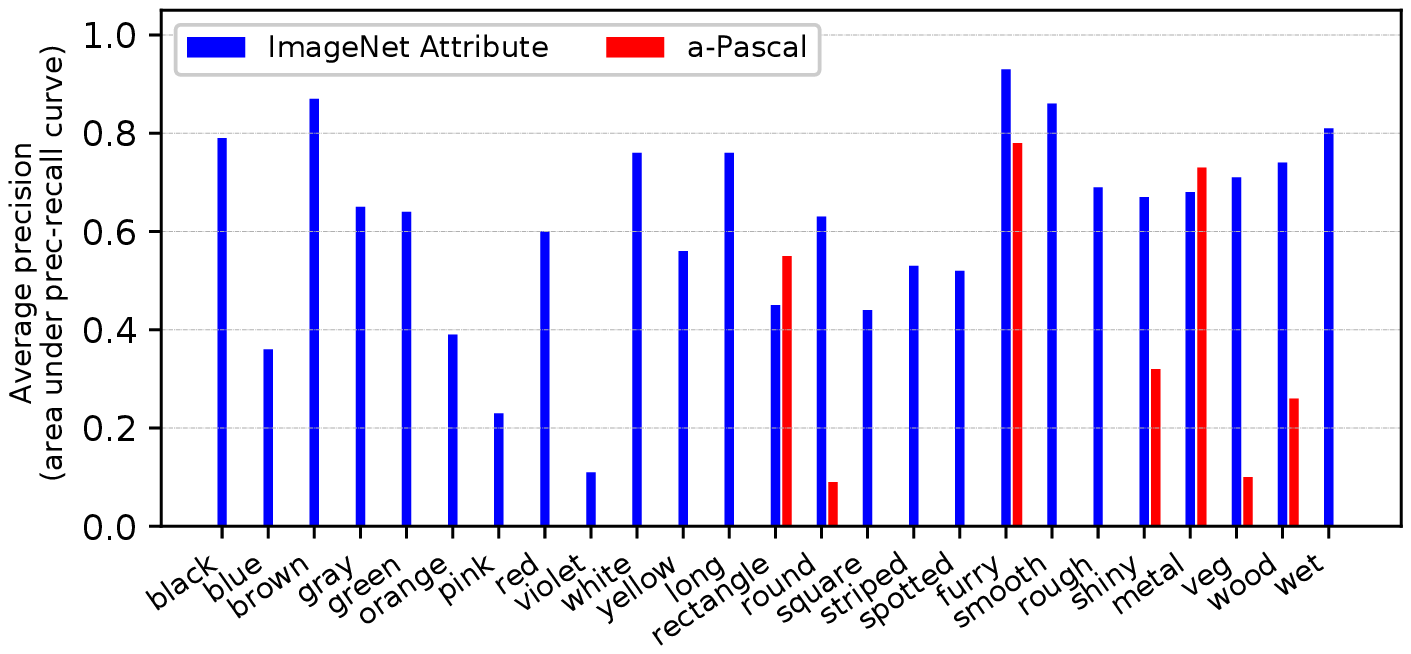}
\caption{Class-wise Average-Precision on the ImageNet Attribute and a-Pascal datasets.
}
\label{fig:IA_AP}
\end{figure}

    We evaluate the classification performance on the test sets of the ImageNet Attribute and a-Pascal datasets.
The ambiguous labels in the ImageNet Attribute dataset are converted to positive labels for computing the precision and recall.
The P-R and ROC curves for the test datasets are shown in Fig. \ref{fig:classification_eval}. 
DAN achieves an overall mAP score of 0.74 in micro mode and a ROC-AUC score of 0.90 for the ImageNet Attribute dataset. 
According to the group mAP, texture performs the best, followed by color, shape, and pattern.
Images from the pattern classes \textit{striped} and \textit{spotted} vary greatly in appearance, as can be seen in Fig. \ref{fig:striped_example}. 
The fraction of ambiguous samples for these two classes is also high, 68\% for \textit{striped}, and 76\% for \textit{spotted}. 
This makes the pattern category difficult for our classifier. 
For the selected shape and texture attributes from the a-Pascal dataset, DAN performs relatively poor by scoring a micro mode mAP of 0.45 and a ROC-AUC score of 0.75.
The AP scores of individual classes are investigated further in Fig.~\ref{fig:IA_AP}.

For the ImageNet Attribute dataset, the AP scores are consistent across all texture classes, resulting in a good mAP score for texture. 
Among the color classes, the two lowest scores are obtained for \textit{pink} and \textit{violet}.
This can be attributed to the lack of sufficient training examples.
There are only $78$ samples for \textit{pink} and $34$ samples for \textit{violet} out of $5760$ training samples. 
\begin{figure}[t!]
\centering
\includegraphics[width=0.98\linewidth]{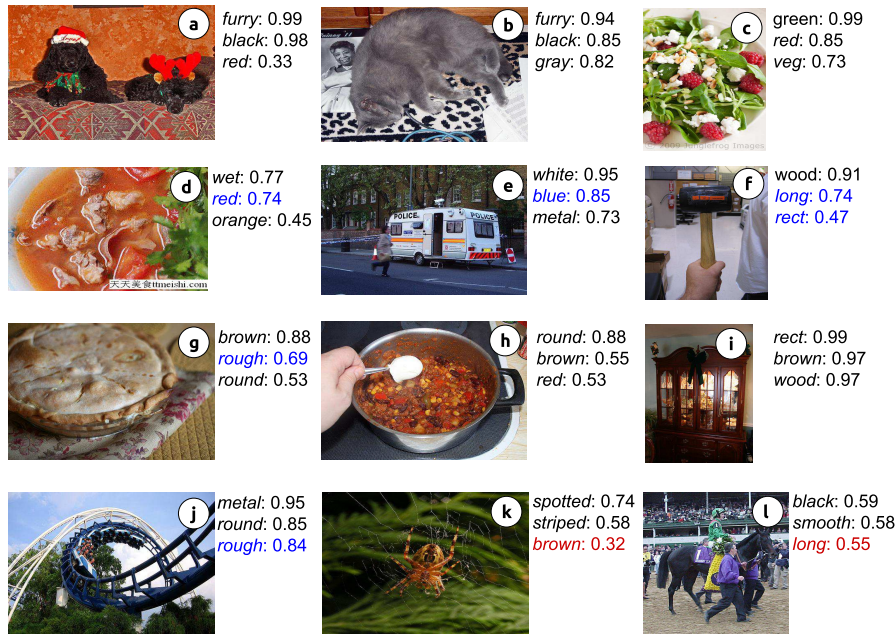}
\caption{Attribute predictions by DAN on sample images from the ImageNet Attribute dataset. For each image, the attributes with top 3 scores are reported. Correct and incorrect predictions are highlighted in black and red respectively. Blue signifies that the prediction is correct, but is considered wrong due to faulty annotations.}
\label{fig:sample_op}
\end{figure}
Fig. \ref{fig:IA_AP} also shows the performance of the individual classes from the a-Pascal dataset. 
Though the overall mAP score for this dataset is poor, the class-wise AP scores for the attributes \textit{rectangle}, \textit{furry} and \textit{metal} are at par with the ImageNet Attribute dataset.
For the rest of the attributes, the object types differ from the training dataset.
For example, the attribute \textit{round} in a-Pascal describes part of objects (wheel of \textit{car}, logo on \textit{train}), whereas in ImageNet Attribute it describes the shape of the objects as a whole.
The \textit{vegetation} attribute in a-Pascal dataset describes \textit{pottedplant} in indoor setup, which is very different from the training samples that describe plants in the wild. 
This affects the classification performance for these attributes.

Fig. \ref{fig:sample_op} shows the attribute classification results for a few sample images from the ImageNet Attribute dataset. The figure shows both correct and incorrect predictions. For example, all the top 3 predicted attributes for the sample images (a,b,c,h,i) are correct. The true label \textit{orange} is incorrectly predicted as \textit{brown} in sample image (k). For sample (l), the attribute \textit{violet} is not predicted. The ImageNet Attribute dataset contains a number of samples, where some object attributes are not annotated. Samples (d,e,f,g,j) highlight such attribute predictions in blue.

\paragraph{\textbf{Evaluation on cropped test data:}}
\setlength{\tabcolsep}{4pt}
\begin{table}[b!]
\centering
\caption{Micro and macro mode mAP on the ImageNet Attribute dataset for un-cropped and cropped test images.}
\label{tab:map_micro_cropped_IA}
\resizebox{0.8\textwidth}{!}{
\begin{tabular}{ccccccc}
\toprule
&mAP&All&Color&Shape&Pattern&Texture\\
\midrule
Un-cropped&micro&$0.74$&$0.73$&$0.62$&$0.52$&$0.81$\\ 
Un-cropped&macro&$0.61$&$0.54$&$0.56$&$0.52$&$0.75$\\ 
\midrule
Cropped&micro&$0.78$&$0.77$&$0.68$&$0.59$&$0.83$\\
Cropped&macro&$0.66$&$0.58$&$0.64$&$0.59$&$0.80$\\
\bottomrule
\end{tabular}
}
\end{table}
\setlength{\tabcolsep}{1.4pt}
The images in the ImageNet Attribute dataset are annotated at object level. 
Hence any predicted attribute that corresponds to the background is considered false in the evaluation, even if it is correct for the background.
To understand the true performance, we also crop the test images to the object bounding boxes with an added margin of 10\%.
Table~\ref{tab:map_micro_cropped_IA} reports the micro and macro mode mAP for the un-cropped and the cropped images from the ImageNet Attribute test set.
The performance consistently improves for all attribute categories by $2-7\%$ in micro mode and $4-8\%$ in macro mode compared to using the entire image.
The lower macro-mAP scores in both cases suggest that the micro-mAP scores are affected by the class imbalance of the dataset.
 
	\paragraph{\textbf{Comparison with state-of-the-art methods:}}
	We compare the performance of the Deep Attribute Network to two state-of-the-art approaches - Russakovsky et al. \cite{Russakovsky2012} and Razavian et al. \cite{razavian2014cnn}.
	Russakovsky et al.~\cite{Russakovsky2012} perform low-level object attribute classification by using handcrafted features and SVM classifiers on the ImageNet Attribute dataset.
	Razavian et al.~\cite{razavian2014cnn}, on the other hand, perform high-level object attribute classification using features extracted from a ConvNet.
	
	We replicate the approach of Razavian et al.~\cite{razavian2014cnn} for low-level attribute classification and design an experiment named VGG-16+SVM.
	We use a pre-trained VGG-16 network as a feature extractor and train four SVMs for the four attribute groups with these features on our training set.
	Similar to Razavian et al.'s set up~\cite{razavian2014cnn}, we use the features from the first fully connected layer of the ConvNet, follow L2 normalization on the features, and use Radial Basis Function (RBF) kernel in the SVMs.
	The VGG-16+SVM experiment acts as a bridge between the method of Russakovsky et al. involving handcrafted features and the ConvNet based approach proposed in this paper.
	
	We also test our proposed method by using the ResNet-50 ConvNet \cite{he2016deep} in place of VGG-16 in order to verify the validity of our approach.
	We replace the last layer of ResNet with a new fully connected layer containing 25 output nodes and train it for 16 epochs on our training set.
	All other training parameters are same as DAN.
	This network is referred to as DAN-ResNet.
	
\setlength{\tabcolsep}{4pt}
\begin{table}[t!]
\centering
\caption{Comparison of ROC-AUC scores of the proposed approaches with state-of-the-art methods on 20 selected attributes from the ImageNet Attribute test set.
Classes \textit{blue}, \textit{violet}, \textit{pink}, \textit{square} and \textit{vegetation} are excluded for a fair comparison with \cite{Russakovsky2012}.
}
\label{tab:DAN_results_comparison1}
\resizebox{0.9\textwidth}{!}{%
\begin{tabular}{@{}ccccc@{}}
\toprule
        & Russakovsky et al. \cite{Russakovsky2012} & VGG-16+SVM & DAN  & DAN-ResNet \\
\midrule
Color   & 0.87               & 0.81       & 0.90 & 0.91             \\
Shape   & 0.83               & 0.77       & 0.87 & 0.87             \\
Pattern & 0.63               & 0.60       & 0.81 & 0.82             \\
Texture & 0.77               & 0.84       & 0.92 & 0.91            \\
\bottomrule
\end{tabular}%
}
\end{table}
\setlength{\tabcolsep}{1.4pt}

	Russakovsky et al.~\cite{Russakovsky2012} exclude the ambiguous records, and also the classes \textit{blue}, \textit{violet}, \textit{pink}, \textit{square} and \textit{vegetation} due to insufficient training samples ($<$ 75 positive samples).
	In order to conduct a fair comparison, we also exclude these classes in the following experiment.
	As our models jointly learn all attributes, filtering ambiguous records would remove records with even a single ambiguous class label ($\approx96\%$ records).
	Hence, the ambiguous samples are not excluded for DAN and DAN-ResNet.
	For VGG-16+SVM, the ambiguous labels are converted to positive labels.
	The ROC-AUC scores of Russakovsky et al.'s method \cite{Russakovsky2012}, VGG-16+SVM, DAN and DAN-ResNet on the ImageNet Attribute test set are shown in Table \ref{tab:DAN_results_comparison1}.
	The networks based on our proposed approach (DAN and DAN-ResNet) outperform Russakovsky et al.'s approach and VGG-16+SVM by $3-18\%$ and $8-21\%$ respectively.
	Particularly, the pattern attribute group shows a major improvement.
	The results indicate the advantages of learned features over handcrafted features.
	The relatively poor performance of VGG-16+SVM suggests that features extracted from object classification ConvNet are not sufficient for the attribute classification task and need further fine-tuning.

\setlength{\tabcolsep}{4pt}
\begin{table}[t!]
\centering
\caption{Comparison of micro-mode mAP scores of the proposed approaches with state-of-the-art methods on the ImageNet Attribute test set. 
}
\label{tab:DAN_results_comparison2}
\resizebox{0.75\textwidth}{!}{%
\begin{tabular}{@{}ccccccc@{}}
\toprule
                                 & \multicolumn{2}{c}{VGG-16+SVM} & \multicolumn{2}{c}{DAN} & \multicolumn{2}{c}{DAN-ResNet} \\
\midrule
\#attributes & 20             & 25            & 20         & 25         & 20                & 25               \\
\midrule
Color                            & 0.52           & 0.51          & 0.75       & 0.73       & 0.75              & 0.74             \\
Shape                            & 0.45           & 0.41          & 0.66       & 0.62       & 0.66              & 0.63             \\
Pattern                          & 0.18           & 0.18          & 0.52       & 0.52       & 0.53              & 0.53             \\
Texture                          & 0.64           & 0.63          & 0.82       & 0.81       & 0.79              & 0.77  \\
\bottomrule          
\end{tabular}%
}
\end{table}
\setlength{\tabcolsep}{1.4pt}
	
In a multi-label setup, ROC-AUC score gets biased by the naturally occurring large number of true negatives, as in the case of the ImageNet Attribute dataset.
The mAP metric does not suffer from this bias and hence better represents a model's performance. 
Table \ref{tab:DAN_results_comparison2} shows the micro-mode mAP scores of VGG-16+SVM, DAN and DAN-ResNet on the ImageNet Attribute test set. Russakovsky et al. only report the ROC-AUC score~\cite{Russakovsky2012}.
As their trained models are not available, the mAP scores could not be computed for the same.
The mAP scores are calculated for the 20 attributes as in Table \ref{tab:DAN_results_comparison1} and also for all 25 attributes.
When the 5 difficult classes (\textit{blue}, \textit{violet}, \textit{pink}, \textit{square} and \textit{vegetation}) are included, the performance decreases consistently.
The similar performance of DAN and DAN-ResNet in both Table~\ref{tab:DAN_results_comparison1} and~\ref{tab:DAN_results_comparison2} demonstrates the validity of our approach.

	\begin{figure}[b!]
	\centering
	\includegraphics[width=\linewidth]{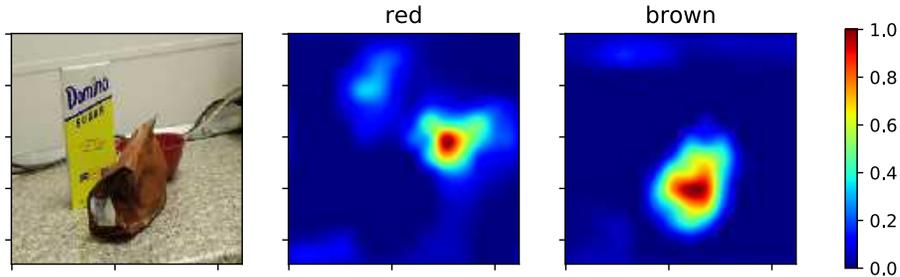}
	\caption{Attention maps generated using Excitation Backprop~\cite{zhang2016top} for attributes \textit{red} and \textit{brown} for a sample input image (shown in the left most column).} 
	\label{fig:att_maps}
	\end{figure}
	
\paragraph{\textbf{Object localization using attributes:}}
It is interesting to see if objects can be localized by using their attributes.
In the past, attention maps generated from object classification ConvNets have been used for locating objects~\cite{zhang2016top}.
An attention map shows the highly activated regions of the image which are responsible for a particular classification decision.
As part of a preliminary study on attribute localization, we use an algorithm called Excitation Backprop\cite{zhang2016top} for generating attention maps from DAN. To do so, we provide the attribute to be localized as a feedback signal to the network.
Using this approach, attention maps are generated for a sample image for the attributes \textit{red} and \textit{brown}, as shown in Fig.~\ref{fig:att_maps}. It can be seen that the location of the activation changes depending on the feedback attribute. The red bowl in the input image, though being partially occluded, can be detected using the attribute \textit{red}. This result hints at the promise of using attributes for localizing objects and we will investigate this approach further in our future work.

	\section{Conclusion}
	\label{sec:conclusion}
In this paper, we have presented a convolutional neural network for low-level object attribute classification. 
The novelty of this method is its simple architecture.
The network is trained from an object classification ConvNet using transfer learning.
It learns 25 attributes belonging to the attribute groups \textit{color}, \textit{shape}, \textit{pattern} and \textit{texture}. 
Experimental results show that our method outperforms state-of-the-art methods.
In the future, we intend to extend this model for attribute localization, which can be helpful in a cluttered or unknown environment, and for unknown objects. 
The attribute specific attention maps shown in section \ref{sec:exp} indicates the plausibility of this future work.
	
	\bibliographystyle{splncs03}
	\bibliography{egbib1}

\end{document}